\pdfoutput=1

\documentclass[11pt]{article}

\usepackage{naacl2021}

\usepackage{times}
\usepackage{latexsym}
\usepackage{multirow, tabularx, booktabs, arydshln}
\usepackage{graphicx}  
\usepackage{subfig}
\usepackage{amssymb}
\usepackage{amsmath} 
\usepackage{multirow}
\usepackage{url}
\usepackage{footnote}
\usepackage{xcolor}
\usepackage{paralist}
\usepackage{enumitem}

\usepackage[T1]{fontenc}

\usepackage[utf8]{inputenc}

\usepackage{microtype}

\DeclareMathOperator*{\argmax}{arg\,max}

\title{Fast and Scalable  Dialogue  State  Tracking  with  Explicit Modular Decomposition}

\author{Dingmin Wang$^{\clubsuit}$, ~ Chenghua Lin$^{\spadesuit}$, ~Qi Liu$^{\clubsuit}$, ~ Kam-Fai Wong$^{\diamondsuit}$\\
$^{\clubsuit}$Department of Computer Science, University of Oxford, UK\\
$^{\spadesuit}$ Department of Computer Science, The University of Sheffield, UK\\
$^{\diamondsuit}$ The Chinese University of Hong Kong, Hong Kong SAR\\
\texttt{\{dingmin.wang, qi.liu\}@cs.ox.ac.uk}~~~~\texttt{c.lin@shef.ac.uk}\\~~~~\texttt{kfwong@se.cuhk.edu.hk}\\
}

\begin{document}
\maketitle
\begin{abstract}
We present a fast and scalable architecture called Explicit Modular Decomposition (EMD), in which we incorporate both classification-based and extraction-based methods and design four modules (for classification and sequence labelling) to jointly extract dialogue states. Experimental results based on the MultiWoz 2.0 dataset validates the superiority of our proposed model in terms of both complexity and scalability when compared to the state-of-the-art methods, especially in the scenario of multi-domain dialogues entangled with many turns of utterances.

\end{abstract}

\section{Introduction}
\begin{figure}[htbp]
  \centering
  \includegraphics[width=0.48\textwidth]{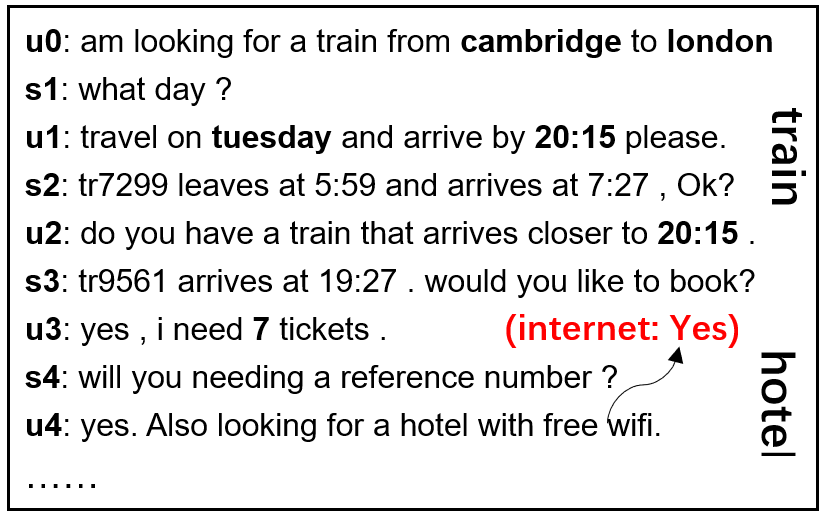}
  \caption{A multi-domain dialogue example extracted from MultiWoz 2.0. The \textbf{S-type} slot values are marked in bold and the arrow points to a pair of \textbf{C-type} slots and its corresponding value. The domain discussed changes from ``train'' to ``hotel'' at the fourth turn. Refer to Section~\ref{sec:model} for the definitions of \textbf{C-type} and \textbf{S-type}. \label{fig:example}}
  \vspace{-0.4cm}
\end{figure}

Dialogue state tracking (DST), responsible for extracting user goals/intentions from dialogues, is a core component in task-oriented dialogue systems~\cite{young2013pomdp}. A dialogue state is commonly represented as a (\textsc{domain}, \textsc{slot type}, \textsc{slot value}) triplet, e.g., (hotel, people, 3). We show an illustrated example of a multi-domain dialogue in Figure~\ref{fig:example}, which involves two domains, i.e., \textsc{train} and \textsc{hotel}.

Previous approaches for DST usually fall into the following four categories: (1) adopt encoder-decoder models to generates states ~\cite{DBLP:conf/acl/KimYKL20,ren2019scalable,li-etal-2019-dual,DBLP:conf/acl/LeeLK19,DBLP:conf/acl/WuMHXSF19} ; (2) cast DST as a multi-label classification task when a full candidate-value list is available~\cite{DBLP:conf/acl/ShanLZMFNZ20, DBLP:conf/acl/RamadanBG18,zhong2018global,DBLP:conf/emnlp/RenXCY18}; (3) employ span-based methods to directly extract the states~\cite{DBLP:conf/interspeech/ChaoL19,DBLP:conf/sigdial/GaoSACH19}; and (4) combine both classification-based and span-based methods to jointly complete the dialogue state extraction~\cite{DBLP:journals/corr/abs-1910-03544}.

The most related work to ours is DS-DST~\cite{DBLP:journals/corr/abs-1910-03544}, a joint model which highlights the problem that using classification-based or span-based approach alone is insufficient to cover all cases of DST in the task-oriented dialogue. While DS-DST has achieved some promising result on dialogue state tracking and demonstrated the utility of combining these two types of methods, some problems still remain unaddressed. On one hand, since the model is conditioned on domain-slot pairs, the computational complexity is not constant and will grow as the number of  domains and slots involved in dialogues increases. 
To be more specific, if there are $1000$ domain-slot pairs, the model needs to run $1000$ times to obtain the expected dialogue states for the current turn at each time, which is a huge computational overhead. On the other hand, previous works usually directly concatenate the history content and the current utterance as input, which is difficult to scale in the multi-turn scenarios, especially when the number of turns of a dialogue is large. Furthermore, we observe that generative approaches may generate some  \textit{domain outlier}\footnote{We refer a predicted result as ``domain outlier'' when slot types are out of the domain pertaining to current utterances.} triplets due to lack of domain constraints. 

To tackle these issues, we propose a fast and scalable method called EMD, where we decompose DST into three classification modules and one sequence labeling module to jointly extract the dialogue states. The benefits of our approach are summarised below:

\begin{itemize}[nosep,leftmargin=1em,labelwidth=*,align=left]  
   \item  \textit{Efficient}: Different to the previous work, we employ a sequence labeling approach to directly annotate the domain-slot values in the utterance instead of iterating over all domain-slot pairs one by one, and thus greatly reduce the model complexity. 
   
   \item \textit{Constrained output}: To effectively model the relationship between the predicted domain and its associated slots, as well as to reduce the occurrence of \textit{domain outlier} results, we propose a list-wise global ranking approach which uses Kullback-Leibler divergence to formulate the training objective. 
   
   \item \textit{Scalable}: Based on turn-level utterances rather than the whole history dialogue content, our proposed model offers better scalability, especially in tackling dialogues with multiple turns. Additionally, we employ a correction module to handle the changes of the states as the dialogue proceeds.
 \end{itemize}  

\begin{figure*}[htb]
  \centering
  \includegraphics[width=0.85\textwidth]{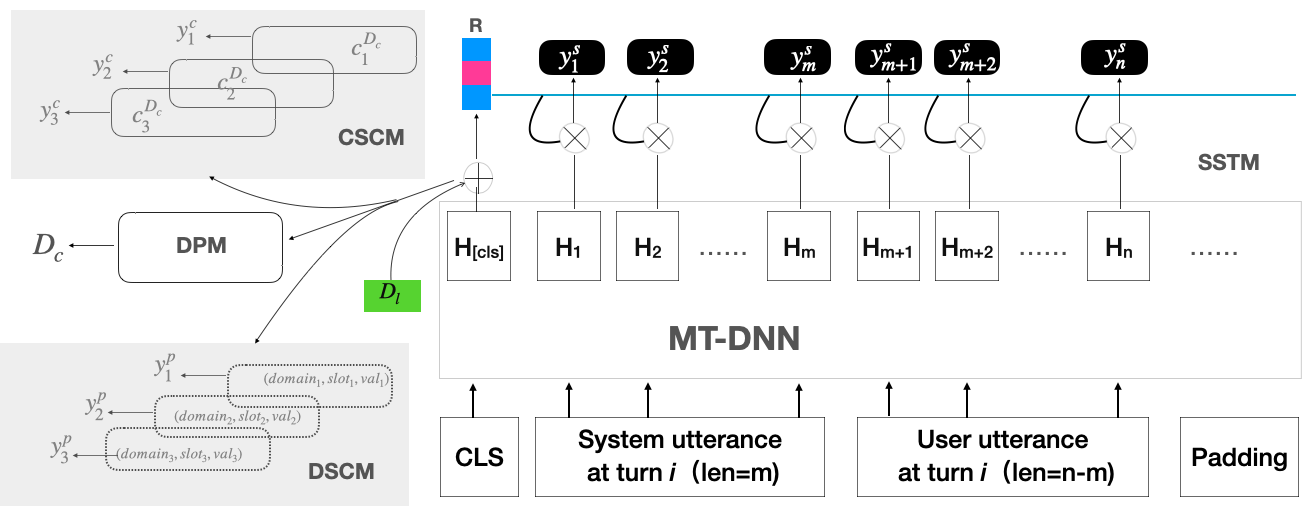}
  \caption{Our neural model architecture, which includes DPM for the domain prediction, whose output is the predicted domain, $D_c$. $D_l$ denotes the domain at the previous turn. CSCM for the three classification of the domain-associated \textbf{C-type} slots, in which $c_i^{D_c}$ denotes one of \textbf{C-type} slots in $D_c$,  and SSTM for tagging \textbf{S-type} slots in the given input, where tagging results are in IOB format; DSCM is for deciding whether to remove outdated states from the history state set. $y_i^p \in \{\text{yes, no}\}$, $y_i^c \in \{\text{yes, no, don't  care}\}$ and $y_i^s \in \{O\} \bigcup$ \{all $\textbf{S-type}$ slots\}.
\vspace{-0.1cm}
\label{fig:achitecture}}
\end{figure*}
\section{Our Proposed Model}\label{sec:model}
Formally, a multi-turn dialogue is represented as $T = \{(s_1, u_1, d_1), (s_2, u_2, d_2), \cdots, (s_n, u_n, d_{n})\}$, $d_i \in D$,  where $s_i$, $u_i$ and $d_i$ refer to the system utterance, the user utterance, and the domain at turn $i$, respectively\footnote{We assume that the turn-level utterances only contain one domain, and the Multiwoz 2.0 dataset we use in this paper also conforms to this assumption.}, and $D$ represents the set of all domains in the training dataset. The overall architecture of our  model is shown in Figure~\ref{fig:achitecture}. 

In our proposed model, we choose MT-DNN~\cite{DBLP:conf/acl/LiuHCG19},  pretrained model which has the same architecture as BERT but trained on multiple
GLUE tasks~\cite{DBLP:conf/iclr/WangSMHLB19}. MT-DNN has been shown to be a better contextual feature extractor for downstream NLP tasks. Given dialogue utterances as input, we represent the output of MT-DNN as $\{H_{[CLS]}, H_1, H_2, \cdots, H_n\}$, where $n$ is the length of the concatenation of the system and user utterances. As a sentence-level representation,  $H_{[CLS]}$ is expected to encode the information of the whole input sequence~\cite{DBLP:conf/naacl/DevlinCLT19, DBLP:conf/acl/LiuHCG19}. Based on these contextual representations, we predict the domain (see \S\ref{ssec:dpm}) and belief states (see \S\ref{ssec:stm} and \S\ref{ssec:cscm}).

Figure~\ref{fig:example} shows a typical multi-domain dialogue example, from which we can observe that some slot values can be directly found from utterances (e.g. \texttt{cambridge} and \texttt{london}), while other slot values are implicit which are more challenging to discover, e.g., requiring classification to infer the values (e.g. \texttt{internet:Yes}). We divide slots into two categories that are handled by two two separate modules: \textbf{S-type} slots whose values could be extracted from dialogue utterances, and \textbf{C-type} slots whose values do not appear in utterances and are chosen from one of the three values $\{\text{yes, no, don't  care}\}$. 

\subsection{Domain Prediction Module ({\small DPM})}\label{ssec:dpm}
In a multi-domain dialogue, the target domain may change as the dialogue proceeds. Different from some previous works~\cite{DBLP:journals/corr/abs-1902-10909, DBLP:journals/corr/abs-1907-02884}, which directly use the first hidden state ($H_{[CLS]}$), in our model, apart from $H_{[CLS]}$, we additionally incorporate $D_{l}$, the domain result of the last turn into the our domain prediction module. The rationale behind is that when the domain of current utterances is not explicit, $D_{l}$ can provide useful reference information for domain identification. Formally, the domain is predicted as:
\begin{equation}
    y^{d} = \text{softmax}(W^{d} [H_{[CLS]};E(D_{l})])
    \vspace{-0.4cm}
\end{equation}
\begin{equation}
        D_c = \argmax(y^{d}), D_c\in D
\end{equation}
where $;$ denotes the concatenation operation and $E(\cdot)$ embeds a word into a distributed representation using fixed MT-DNN~\cite{DBLP:conf/acl/LiuHCG19}. $D_c$ is the predicted domain result.

\subsection{\textbf{S-type} Slots Tagging Module ({\small SSTM})}\label{ssec:stm}
\paragraph{Domain-slot-matching constraints $R$} To prevent our model from predicting some slots not belonging to the current domain, we generate a domain constrained contextual record $R \in \mathbb{R}^{1 \times (s+1)}$, where $s$ is number of \textbf{S-type} slots of all domains\footnote{We add a \textit{[EMPTY]}, the value of which is expected to be $1$ when there is no slot needed to be predicted. In particular, we consider the ``don't care'' as a special case in which the corresponding slot is considered not to be predicted.}. Concretely speaking, $R$ is a distribution over all \textbf{S-type} slots and \textit{[EMPTY]} using
\begin{equation}
    R = \text{softmax}(W^{R} [H_{[CLS]};E(D_{l}])
\end{equation}
In particular, $L_{R}$, the loss for $R$ is defined as the Kullback-Leibler (KL) divergence between $Div(R^{real}||R)$, where distribution $R^{real}$ from the ground truth is computed as follows:
\begin{itemize}[nosep,leftmargin=1em,labelwidth=*,align=left]
    \item If there is no slot required to be predicted, $R^{real}_{[EMPTY]}$ receives a probability mass of $1$ for the special slot \textit{[EMPTY]}.
    \item If the number of slots needed to be predicted is $k(\geq 1)$, then corresponding $k$ slot positions receive an equal probability mass of $1/k$.
\end{itemize}
Next, we employ a sequence labeling approach to directly annotate the domain-slot values in the utterance instead of iterating over all domain-slot pairs one by one. Specifically, to tag \textbf{S-type} slots of the given input, we feed the final hidden states of $H_1, H_2, \cdots, H_n$ into a softmax layer to classify all the \textbf{S-type} slots,
\begin{equation}
    y_i^{s} = \text{softmax}(W^{s}H_i ), i \in [1, 2, \cdots, N]
\end{equation}
Instead of directly predicting \textbf{S-type} slot results based on $y_i^{s}$, we introduce a domain-slot-matching constraint $R$, which helps avoid generating \textbf{S-type} slots that do not belong to the predicted domain. The multiplication operation is given below,
\begin{equation}
   \hat{ y_i^{s}} = R  \odot  y_i^{s}
\end{equation}
 where $\odot$ is the element-wise multiplication.

\subsection{C-type Slots Classification Module~({\small CSCM})}\label{ssec:cscm}
Given the currently predicted domain result $D_c$, we build a set $C_{D_c}$ which contains all \textbf{C-type} slots from all domains $D$. If $C_{D_c}$ is empty, it indicates that there is no \textbf{C-type} slot needed to be predicted in the current domain.  Otherwise, we classify each slot $c_i^{D_c}$ in $C_{D}$ into one of the following following categories, i.e., $\text{\{yes, no, don't care\}}$, with the classification function below.
\begin{equation}
    y^{c} = \text{softmax}(W^{c}[E(c_i^{D_c}); E(D_l); H_{[CLS]}])
\end{equation}

\subsection{Dialogue  State  Correction  Module~({\small DSCM})}\label{ssec:dscm}
Previous models such as TRADE~\cite{DBLP:conf/acl/WuMHXSF19} and  COMER~\cite{ren2019scalable} requires that all dialogue states need to be predicted from scratch at each turn, including those dialogue states that have already been predicted at previous turns. This poses a big challenge to the model in terms of scalability, especially when the number of dialogue turns increases. Conversely, the input of our model consists of the system utterance and the user utterance at the current turn, so our model only outputs the estimates of the dialogue states for the current turn, and the previous dialogues are directly included where no re-prediction is needed.


However, there is an issue with direct inclusion of previously predicted results in that some states may need to be updated or removed as the dialogue proceeds.  For example, a user firstly looks for a hotel located in the center area, then a state (hotel, area, center) is estimated. Subsequently, the user utters a specified hotel name, e.g. ``\textit{I wanna the King House}'', then the previous state (hotel, area, center) is outdated and should be removed. To this end, we design the dialogue state correction module to update previously predicted results in order to improve the precision of the outputted dialogues states at each turn. Similar to the C-type classification module, we cast this situation as a classification task, and for each triple tuple $p$ from the previous dialogue states, the classifier is formulated as
\begin{equation}
    y^{p} = \text{sigmoid}(W^{p}[\hat{p}; E(D_l); H_{[CLS]}])
\end{equation}
Here each item in $p$ is embedded using $E(\cdot)$ and $\hat{p}$ is the embedding sum of the three items in $p$.

During training, we use cross entropy loss for $y^{d}$, $y^{c}$, $y^{s}$ and $y^{p}$, which are represented as $L_{y^{d}}$, $L_{y^{c}}$, $L_{y^{s}}$ and $L_{y^{p}}$, respectively. The loss for $R$ (denoted as $L_{R}$) is defined as Kullback-Leibler (KL) divergence between $R^{real} $ and $R$ (i.e, $\text{KL}(R^{real}||R)$).
All parameters are jointly trained by minimizing the weighted-sum of five losses ($\alpha$, $\beta$, $\gamma$, $\theta$, $\epsilon$ are hyper-parameters),
 \begin{equation}
     \text{Loss} = \alpha L_{y^{d}} + \beta L_{y^{c}}  + \gamma L_{y^{s}} + \theta L_{y^{p}} + \epsilon  L_{R}
 \end{equation}
 
\subsection{Analysis of model complexity}\label{ssec:mc}
Table~\ref{tab:itc} reports the Inference Time Complexity~(ITC) proposed by~\cite{ren2019scalable}, which is used to measure the model complexity. ITC calculates how many times inference must be performed to complete a
prediction of the belief state in a dialogue turn. By comparison, we can observe that our model achieves the lowest complexity, $\mathcal{O}(1)$, attributed to the modular decomposition and the usage of the sequence label based model. 

\begin{table}[htp]
\centering \small
   \begin{tabular}{lccc}
        \toprule
         Model &  ITC \\
         \hline 
        DS-DST \cite{DBLP:journals/corr/abs-1910-03544} &  $\mathcal{O}(n)$ \\
        SOM-DST \cite{DBLP:conf/acl/KimYKL20}  & $\mathcal{O}(n)$\\
        SUMBT \cite{DBLP:conf/acl/LeeLK19} & $\mathcal{O}(mn)$ \\ 
        GLAD \cite{zhong2018global}  & $\mathcal{O}(mn)$ \\
        COMER~\cite{ren2019scalable}n& $\mathcal{O}(n)$ \\
        TRADE~\cite{DBLP:conf/acl/WuMHXSF19} & $\mathcal{O}(n)$ \\
        \hline 
        EMD & $\mathcal{O}(1)$ \\
        \bottomrule
    \end{tabular}
 
  \caption{Inference Time Complexity~(ITC) proposed in~~\cite{ren2019scalable}, m is the number of values in a pre-defined ontology list
and n is the number of slots. Note that the ITC reported refers to the worst scenarios. \label{tab:itc}}
\vspace{-0.7cm}
\end{table}

\section{Experimental Setup}\label{ssec:exp}
\begin{figure*}[htb]
  \centering
  \includegraphics[width=0.99\textwidth]{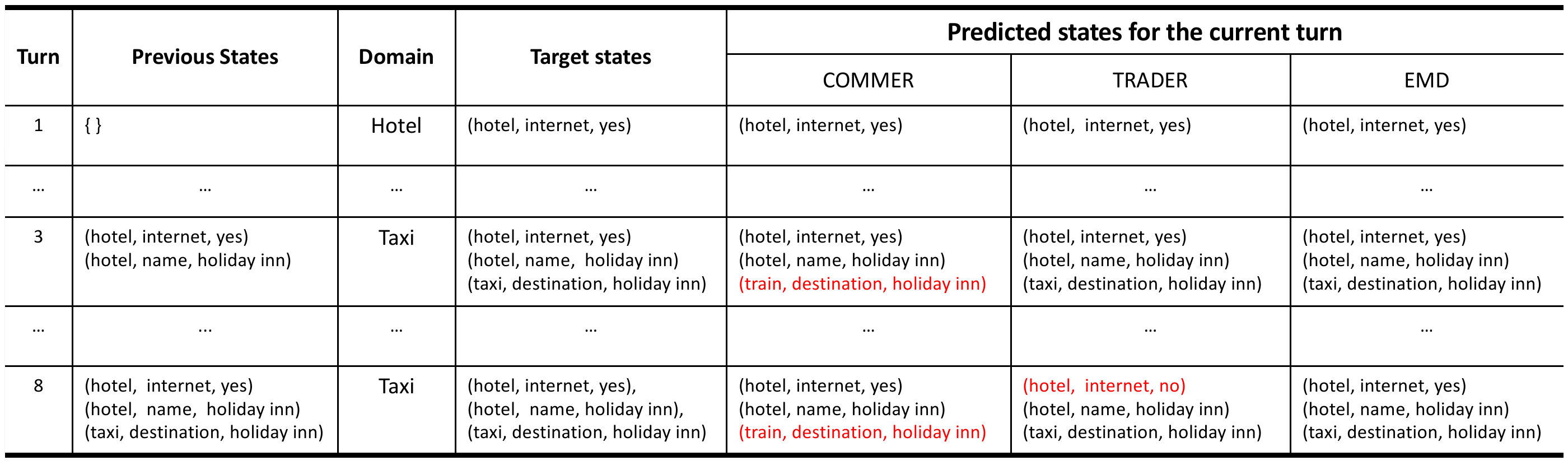}
  \caption{Case study of predicated states by our model and two baselines. Erroneous states are highlighted in red. \label{fig:casestudy}}
\end{figure*}
\subsection{Setup}

\paragraph{Dataset} We evaluate our model performance based on the  MultiWoZ 2.0 dataset~\cite{DBLP:conf/emnlp/BudzianowskiWTC18}, which contains  $10,000$ dialogues of $7$ domains and $35$ domain-slot pairs. Detailed dataset statistics is summarised in Table~\ref{tab:data}. 

\paragraph{Evaluation metrics} We utilize joint goal accuracy~(JGA)~\cite{henderson2014second} to evaluate the model performance. Joint goal accuracy is the accuracy of the dialogue state of each turn and a dialogue state is regarded as correct only if all the values of slots are correctly predicted. 

\paragraph{Implementation details} The hyper-parameters of our model go as follows: both the embedding and the hidden size is $1024$; we used a learning rate of 0.0001 with a gradient clip of 2.0, mini-batch SGD with a batch size of $32$, and Adam optimizer~\cite{kingma2014adam} for $50$ epoch training. We set a value of $1$ to the five weighted hyper-parameters:
$\alpha$, $\beta$, $\gamma$, $\theta$, $\epsilon$.
\begin{table}[htp]
\centering \small

  \renewcommand{\arraystretch}{1}
  \setlength{\tabcolsep}{1mm}{
\begin{tabular}{lccc}
        \toprule
         Metric &  Train  & Dev & Test\\
         \hline 
        \# of multi-domain dialogs & 5,459 & 796 & 777\\
        \# of single-domain dialogs & 2,979 & 204 & 223 \\
        \# of total dialogs  & 8,438 & 1,000 & 1,000 \\ 
        Avg. \# turns by dialog & 6.7 & 7.4 & 7.3\\
        \bottomrule 
    \end{tabular}
  }
  \caption{The statistics of the MultiWoZ2.0.\label{tab:data}}

\end{table}

\subsection{Results}
\noindent\textbf{Overall comparison}~~We compare our models against six strong baselines on the multi-domain dataset MultiWoz. Results are reported in Table~\ref{tab:result} based on joint goal accuracy~(JGA). Our model achieves the best performance of $50.18\%$ in the multi-domain testset, while the accuracy achieved in the single-domain is on par with the state-of-the-art results, which demonstrates the superiority of our model. 
\begin{table}[ht]
\centering \small
 \renewcommand{\arraystretch}{1.2}
 \setlength{\tabcolsep}{1mm}{
   \begin{tabular}{lccc}
        \toprule
         Model & $JGA^{s}$ & $JGA^{m}$ & JGA\\
         \hline 
        SOM-DST \cite{DBLP:conf/acl/KimYKL20} & - & - &  \textbf{51.72}\\
        COMER~\cite{ren2019scalable}& 48.62  & 41.21&  45.72\\
        SUMBT \cite{DBLP:conf/acl/LeeLK19} & 46.99  & 39.68 &  42.40\\
        DS-DST \cite{DBLP:journals/corr/abs-1910-03544} & \textbf{51.99}  & 48.69 &  51.01\\
        GLAD \cite{zhong2018global} & 37.19  & 33.76 &  35.58\\
        TRADE~\cite{DBLP:conf/acl/WuMHXSF19}  & 49.57  & 47.01 &  48.62\\ \hline
        EMD & 51.92  & \textbf{50.18} &  51.03 \\
        \bottomrule
    \end{tabular}
  }
  \caption{Experimental results. $JGA^{s}$ represents the accuracy calculated in all single domain dialogues and $JGA^{m}$ refers to all multi-domain dialogues.\label{tab:result}}
  \vspace{-0.4cm}
\end{table}

\paragraph{Analysis of model scalability}
We select $200$ samples from the testing dataset, in which each dialogue has more than $8$ turns of utterances between the system and the user. Then, taking the turn number $6$ as a threshold, we divide the dialogue content into two categories, i.e., \textsc{Cold} and \textsc{Hot}. Utterances with turn numbers lower than $6$ are assigned to the \textsc{Cold} category and those above $6$ to the \textsc{Hot} category. 

\begin{table}[ht]
\centering \small

 \renewcommand{\arraystretch}{1}
 \setlength{\tabcolsep}{1mm}{
   \begin{tabular}{lcc}
        \toprule
         \multirow{2}{*}{Model} & \multicolumn{2}{c}{JGA}\\
         \cline{2-3} 
         & \textsc{Cold} & \textsc{Hot} \\
         \hline 
        SOM-DST \cite{DBLP:conf/acl/KimYKL20} & \textbf{52.21} & 48.92\\
        COMER~\cite{ren2019scalable}&  46.01&  40.72\\
        SUMBT \cite{DBLP:conf/acl/LeeLK19} & 42.51 &  33.99\\
        TRADE~\cite{DBLP:conf/acl/WuMHXSF19}  & 47.98&  46.12\\ \hline
        EMD   & 51.89 & \textbf{51.01} \\
        \bottomrule
    \end{tabular}
  }
  \caption{Experimental results for the analysis of model scalabitiy. The sample size is $200$. \label{tab:scalability}}
  \vspace{-0.2cm}
\end{table}

From Table~\ref{tab:scalability}, we observe that the model performance has a big drop for the four baseline models, but our model achieves a relatively stable performance, achieving 51.01\% in \textsc{Hot} and $51.89\%$ in \textsc{Cold}, respectively. This demonstrates that our model is not only fast in terms of inference speed~(cf. \S\ref{ssec:mc}), but also has a good scalability which can maintain a high accuracy even when the dialogue proceeds into more turns and the input length becomes larger. 

\paragraph{Ablation study}
We conduct two ablation experiments to investigate the impacts of $D_l$ and  $R$. We introduce a metric, called outlierslot ratio (OSR), denoting the proportion of slots predicted by our model that do not belong to the current domain. From Table~\ref{tab:ablation}, we notice that adding $D_l$ improves the domain accuracy, where one possible reason is that some utterances may not have a clear domain attribute, and thus the incorporated previous domain is believed to provide useful guiding information in  domain prediction. Besides, by comparing OSR with and without using $R$, we can observe that using $R$ reduces the proportion of generating slots that do not align to the predicted domain, which further improves the model performance.
\begin{table}[ht]
\centering \small
   \begin{tabular}{lccc}
        \toprule
         Model & Domain Acc. & OSR & JGA \\
         \hline 
          EMD&  95.23  & 44.62  & 51.03 \\
         - $D_l$ &  91.83  & 45.62 & 48.62  \\
         - $R$&  93.19  & 54.83 & 47.23 \\
        \bottomrule
    \end{tabular}
  \caption{Ablation study results. \label{tab:ablation}}
  \vspace{-0.4cm}
\end{table}

\noindent \textbf{Case study}~~To evaluate our proposed model qualitatively, we show an exemplary dialogue and illustrate some generated results by EMD and two baseline models in Figure~\ref{fig:casestudy}. At turn $3$ when the dialogue domain change from \textit{hotel} to \textit{taxi},  COMMER fails to capture the domain information and generates a domain outlier, ``\textit{train}'', which does not conform to the current context. Conversely, dialogue generated by our model always conforms to the domain at the current turn, which may benefit from the incorporation of the domain constrained contextual record $R$. Besides, another observation is that as the dialogue proceeds to the turn $8$ when the history dialogue content accumulates, TRADER makes an incorrect prediction in the hotel-internet slot, which is correctly identified at the turn $1$. One possible reason is that it becomes more challenging for the model to correctly predict all dialogue state from scratch when both the history dialogue content and states involved increase. Instead of repeatedly generating those previously predicted states at each turn, our model only outputs the states for the current turn, and updates previous dialogue states with a separate module.

\section{Conclusion}
In this paper, we propose to decompose DST into multiple submodules to jointly estimate dialogue states. Experimental results based on the MultiWoz 2.0 dataset show that our model not only reduces the model complexity, but also gives high scalability in coping with multi-domain and long task-oriented dialogue scenarios. 
\bibliography{anthology,custom}

\begin{thebibliography}{21}
\expandafter\ifx\csname natexlab\endcsname\relax\def\natexlab#1{#1}\fi

\bibitem[{Budzianowski et~al.(2018)Budzianowski, Wen, Tseng, Casanueva, Ultes,
  Ramadan, and Gasic}]{DBLP:conf/emnlp/BudzianowskiWTC18}
Pawel Budzianowski, Tsung{-}Hsien Wen, Bo{-}Hsiang Tseng, I{\~{n}}igo
  Casanueva, Stefan Ultes, Osman Ramadan, and Milica Gasic. 2018.
\newblock Multiwoz - {A} large-scale multi-domain wizard-of-oz dataset for
  task-oriented dialogue modelling.
\newblock In \emph{Proceedings of the 2018 Conference on Empirical Methods in
  Natural Language Processing, Brussels, Belgium, October 31 - November 4,
  2018}, pages 5016--5026.

\bibitem[{Castellucci et~al.(2019)Castellucci, Bellomaria, Favalli, and
  Romagnoli}]{DBLP:journals/corr/abs-1907-02884}
Giuseppe Castellucci, Valentina Bellomaria, Andrea Favalli, and Raniero
  Romagnoli. 2019.
\newblock \href {http://arxiv.org/abs/1907.02884} {Multi-lingual intent
  detection and slot filling in a joint bert-based model}.
\newblock \emph{CoRR}, abs/1907.02884.

\bibitem[{Chao and Lane(2019)}]{DBLP:conf/interspeech/ChaoL19}
Guan{-}Lin Chao and Ian Lane. 2019.
\newblock {BERT-DST:} scalable end-to-end dialogue state tracking with
  bidirectional encoder representations from transformer.
\newblock In \emph{Interspeech 2019, 20th Annual Conference of the
  International Speech Communication Association, Graz, Austria, 15-19
  September 2019}, pages 1468--1472.

\bibitem[{Chen et~al.(2019)Chen, Zhuo, and
  Wang}]{DBLP:journals/corr/abs-1902-10909}
Qian Chen, Zhu Zhuo, and Wen Wang. 2019.
\newblock \href {http://arxiv.org/abs/1902.10909} {{BERT} for joint intent
  classification and slot filling}.
\newblock \emph{CoRR}, abs/1902.10909.

\bibitem[{Devlin et~al.(2019)Devlin, Chang, Lee, and
  Toutanova}]{DBLP:conf/naacl/DevlinCLT19}
Jacob Devlin, Ming{-}Wei Chang, Kenton Lee, and Kristina Toutanova. 2019.
\newblock {BERT:} pre-training of deep bidirectional transformers for language
  understanding.
\newblock In \emph{Proceedings of the 2019 Conference of the North American
  Chapter of the Association for Computational Linguistics: Human Language
  Technologies, {NAACL-HLT} 2019, Minneapolis, MN, USA, June 2-7, 2019, Volume
  1 (Long and Short Papers)}, pages 4171--4186.

\bibitem[{Gao et~al.(2019)Gao, Sethi, Agarwal, Chung, and
  Hakkani{-}T{\"{u}}r}]{DBLP:conf/sigdial/GaoSACH19}
Shuyang Gao, Abhishek Sethi, Sanchit Agarwal, Tagyoung Chung, and Dilek
  Hakkani{-}T{\"{u}}r. 2019.
\newblock \href {https://doi.org/10.18653/v1/W19-5932} {Dialog state tracking:
  {A} neural reading comprehension approach}.
\newblock In \emph{Proceedings of the 20th Annual SIGdial Meeting on Discourse
  and Dialogue, SIGdial 2019, Stockholm, Sweden, September 11-13, 2019}, pages
  264--273. Association for Computational Linguistics.

\bibitem[{Henderson et~al.(2014)Henderson, Thomson, and
  Williams}]{henderson2014second}
Matthew Henderson, Blaise Thomson, and Jason~D Williams. 2014.
\newblock The second dialog state tracking challenge.
\newblock In \emph{Proceedings of the 15th annual meeting of the special
  interest group on discourse and dialogue (SIGDIAL)}, pages 263--272.

\bibitem[{Kim et~al.(2020)Kim, Yang, Kim, and Lee}]{DBLP:conf/acl/KimYKL20}
Sungdong Kim, Sohee Yang, Gyuwan Kim, and Sang{-}Woo Lee. 2020.
\newblock \href {https://www.aclweb.org/anthology/2020.acl-main.53/} {Efficient
  dialogue state tracking by selectively overwriting memory}.
\newblock In \emph{Proceedings of the 58th Annual Meeting of the Association
  for Computational Linguistics, {ACL} 2020, Online, July 5-10, 2020}, pages
  567--582. Association for Computational Linguistics.

\bibitem[{Kingma and Ba(2014)}]{kingma2014adam}
Diederik~P Kingma and Jimmy Ba. 2014.
\newblock Adam: A method for stochastic optimization.
\newblock \emph{arXiv preprint arXiv:1412.6980}.

\bibitem[{Lee et~al.(2019)Lee, Lee, and Kim}]{DBLP:conf/acl/LeeLK19}
Hwaran Lee, Jinsik Lee, and Tae{-}Yoon Kim. 2019.
\newblock {SUMBT:} slot-utterance matching for universal and scalable belief
  tracking.
\newblock In \emph{Proceedings of the 57th Conference of the Association for
  Computational Linguistics, {ACL} 2019, Florence, Italy, July 28- August 2,
  2019, Volume 1: Long Papers}, pages 5478--5483.

\bibitem[{Li et~al.(2019)Li, Lin, Collinson, Li, and Chen}]{li-etal-2019-dual}
Ruizhe Li, Chenghua Lin, Matthew Collinson, Xiao Li, and Guanyi Chen. 2019.
\newblock \href {https://doi.org/10.18653/v1/K19-1036} {A dual-attention
  hierarchical recurrent neural network for dialogue act classification}.
\newblock In \emph{Proceedings of the 23rd Conference on Computational Natural
  Language Learning (CoNLL)}, pages 383--392, Hong Kong, China. Association for
  Computational Linguistics.

\bibitem[{Liu et~al.(2019)Liu, He, Chen, and Gao}]{DBLP:conf/acl/LiuHCG19}
Xiaodong Liu, Pengcheng He, Weizhu Chen, and Jianfeng Gao. 2019.
\newblock Multi-task deep neural networks for natural language understanding.
\newblock In \emph{Proceedings of the 57th Conference of the Association for
  Computational Linguistics, {ACL} 2019, Florence, Italy, July 28- August 2,
  2019, Volume 1: Long Papers}, pages 4487--4496.

\bibitem[{Ramadan et~al.(2018)Ramadan, Budzianowski, and
  Gasic}]{DBLP:conf/acl/RamadanBG18}
Osman Ramadan, Pawel Budzianowski, and Milica Gasic. 2018.
\newblock Large-scale multi-domain belief tracking with knowledge sharing.
\newblock In \emph{Proceedings of the 56th Annual Meeting of the Association
  for Computational Linguistics, {ACL} 2018, Melbourne, Australia, July 15-20,
  2018, Volume 2: Short Papers}, pages 432--437.

\bibitem[{Ren et~al.(2019)Ren, Ni, and McAuley}]{ren2019scalable}
Liliang Ren, Jianmo Ni, and Julian McAuley. 2019.
\newblock Scalable and accurate dialogue state tracking via hierarchical
  sequence generation.
\newblock In \emph{Proceedings of the 2019 Conference on Empirical Methods in
  Natural Language Processing and the 9th International Joint Conference on
  Natural Language Processing (EMNLP-IJCNLP)}, pages 1876--1885.

\bibitem[{Ren et~al.(2018)Ren, Xie, Chen, and Yu}]{DBLP:conf/emnlp/RenXCY18}
Liliang Ren, Kaige Xie, Lu~Chen, and Kai Yu. 2018.
\newblock Towards universal dialogue state tracking.
\newblock In \emph{Proceedings of the 2018 Conference on Empirical Methods in
  Natural Language Processing, Brussels, Belgium, October 31 - November 4,
  2018}, pages 2780--2786.

\bibitem[{Shan et~al.(2020)Shan, Li, Zhang, Meng, Feng, Niu, and
  Zhou}]{DBLP:conf/acl/ShanLZMFNZ20}
Yong Shan, Zekang Li, Jinchao Zhang, Fandong Meng, Yang Feng, Cheng Niu, and
  Jie Zhou. 2020.
\newblock \href {https://doi.org/10.18653/v1/2020.acl-main.563} {A contextual
  hierarchical attention network with adaptive objective for dialogue state
  tracking}.
\newblock In \emph{Proceedings of the 58th Annual Meeting of the Association
  for Computational Linguistics, {ACL} 2020, Online, July 5-10, 2020}, pages
  6322--6333. Association for Computational Linguistics.

\bibitem[{Wang et~al.(2019)Wang, Singh, Michael, Hill, Levy, and
  Bowman}]{DBLP:conf/iclr/WangSMHLB19}
Alex Wang, Amanpreet Singh, Julian Michael, Felix Hill, Omer Levy, and
  Samuel~R. Bowman. 2019.
\newblock \href {https://openreview.net/forum?id=rJ4km2R5t7} {{GLUE:} {A}
  multi-task benchmark and analysis platform for natural language
  understanding}.
\newblock In \emph{7th International Conference on Learning Representations,
  {ICLR} 2019, New Orleans, LA, USA, May 6-9, 2019}. OpenReview.net.

\bibitem[{Wu et~al.(2019)Wu, Madotto, Hosseini{-}Asl, Xiong, Socher, and
  Fung}]{DBLP:conf/acl/WuMHXSF19}
Chien{-}Sheng Wu, Andrea Madotto, Ehsan Hosseini{-}Asl, Caiming Xiong, Richard
  Socher, and Pascale Fung. 2019.
\newblock Transferable multi-domain state generator for task-oriented dialogue
  systems.
\newblock In \emph{Proceedings of the 57th Conference of the Association for
  Computational Linguistics, {ACL} 2019, Florence, Italy, July 28- August 2,
  2019, Volume 1: Long Papers}, pages 808--819.

\bibitem[{Young et~al.(2013)Young, Ga{\v{s}}i{\'c}, Thomson, and
  Williams}]{young2013pomdp}
Steve Young, Milica Ga{\v{s}}i{\'c}, Blaise Thomson, and Jason~D Williams.
  2013.
\newblock Pomdp-based statistical spoken dialog systems: A review.
\newblock \emph{Proceedings of the IEEE}, 101(5):1160--1179.

\bibitem[{Zhang et~al.(2019)Zhang, Hashimoto, Wu, Wan, Yu, Socher, and
  Xiong}]{DBLP:journals/corr/abs-1910-03544}
Jianguo Zhang, Kazuma Hashimoto, Chien{-}Sheng Wu, Yao Wan, Philip~S. Yu,
  Richard Socher, and Caiming Xiong. 2019.
\newblock \href {http://arxiv.org/abs/1910.03544} {Find or classify? dual
  strategy for slot-value predictions on multi-domain dialog state tracking}.
\newblock \emph{CoRR}, abs/1910.03544.

\bibitem[{Zhong et~al.(2018)Zhong, Xiong, and Socher}]{zhong2018global}
Victor Zhong, Caiming Xiong, and Richard Socher. 2018.
\newblock {Global-Locally Self-Attentive Encoder for Dialogue State Tracking}.
\newblock In \emph{Proceedings of the 56th Annual Meeting of the Association
  for Computational Linguistics (Volume 1: Long Papers)}, pages 1458--1467.

\end{thebibliography}
\bibliographystyle{acl_natbib}

\end{document}